\documentclass[a4paper, 10pt, conference]{ieeeconf}      % Use this line for a4
\usepackage[accsupp]{axessibility} 
\usepackage{FG2021}
\usepackage{hhline}
\usepackage{multirow}
\usepackage{graphicx}
\usepackage{url}

\FGfinalcopy

\IEEEoverridecommandlockouts

\IEEEoverridecommandlockouts\pubid{\makebox[\columnwidth]{978-1-6654-3176-7/21/\$31.00~\copyright{}2021 IEEE \hfill} \hspace{\columnsep}\makebox[\columnwidth]{ }}
\def\FGPaperID{****} % *** Enter the FG2021 Paper ID here

\title{\LARGE \bf
    From Face to Gait: Weakly-Supervised Learning of Gender Information from Walking Patterns
}

\author{\parbox{16cm}{\centering
    {\large Andy Catruna$^{1}$ Adrian Cosma$^{2}$ Ion Emilian Radoi$^{3}$}\\
    {\normalsize
    $^1$andy.eduard.catruna@gmail.com, $^2$cosma.i.adrian@gmail.com, $^3$emilian.radoi@cs.pub.ro\\
    University Politehnica of Bucharest\\
    }
}}

\begin{document}

\ifFGfinal
\thispagestyle{empty}
\pagestyle{empty}
\else
% \author{Anonymous FG2021 submission\\ Paper ID \FGPaperID \\}
\author{}
\pagestyle{plain}
\fi
\maketitle

%%%%%%%%%%%%%%%%%%%%%%%%%%%%%%%%%%%%%%%%%%%%%%%%%%%%%%%%%%%%%%%%%%%%%%%%%%%%%%%%
\begin{abstract}

Obtaining demographics information from video is valuable for a range of real-world applications.
While approaches that leverage facial features for gender inference are very successful in restrained environments, they do not work in most real-world scenarios when the subject is not facing the camera, has the face obstructed or the face is not clear due to distance from the camera or poor resolution. 
We propose a weakly-supervised method for learning gender information of people based on their manner of walking.
We make use of state-of-the art facial analysis models to automatically annotate front-view walking sequences and generalise to unseen angles by leveraging gait-based label propagation. 
Our results show on par or higher performance with facial analysis models with an F1 score of 91\% and the ability to successfully generalise to scenarios in which facial analysis is unfeasible due to subjects not facing the camera or having the face obstructed.
\end{abstract}

%%%%%%%%%%%%%%%%%%%%%%%%%%%%%%%%%%%%%%%%%%%%%%%%%%%%%%%%%%%%%%%%%%%%%%%%%%%%%%%%
\section{INTRODUCTION}
\begin{figure*}[hbt!]
    \centering
    \includegraphics[width=0.65\textwidth]{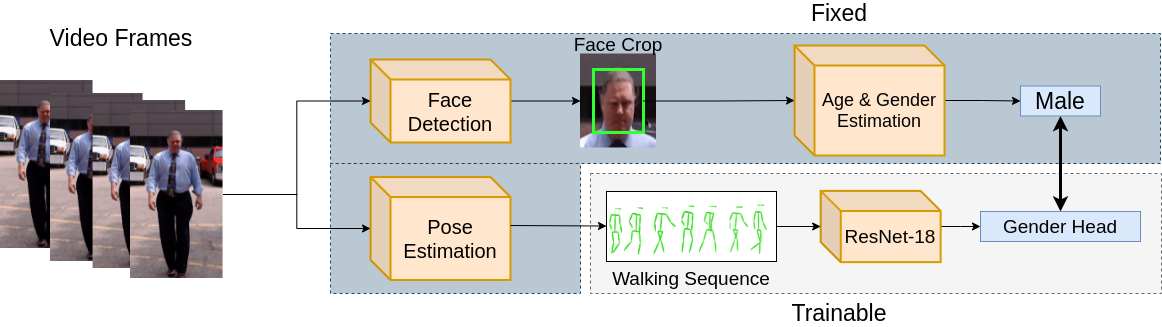}
    \vspace{-2mm}
    \caption{Our method for training gait-based gender identification based on noisy labels given by the facial analysis model. Gender labels are inferred from front-views and are processed by a ResNet-18 architecture which processes skeleton sequences. For the final inference, only skeleton sequences are used.}
    \label{fig:framework}
    \vspace{-4mm}
\end{figure*}

Demographics estimation from video is a long standing research topic in the machine learning community. Gender estimation in images and videos has numerous potential applications in areas such as access control, marketing and surveillance \cite{10.1007/978-3-030-00665-5_69}. Across the years, developing facial analysis models was the primary focus of computer vision research in this area. This led to the construction of several large-scale datasets for modelling a plethora of attributes derived from faces: demographics, emotions, appearance, and person identity. 
Systems based on facial analysis models have matured to the point of deployment in constrained environments \cite{rite-aid}. However, in surveillance scenarios limitations are apparent in situations where the subject is not cooperating (i.e. not facing the camera) or surveillance cameras are placed at a distance thus capturing faces in poor resolution. 

Recent advances in gait analysis systems show that gait holds rich biometric information, being able to reliably determine a person's identity \cite{cosma2021wildgait,9324873}, demographics characteristics \cite{Xu2017} and emotions \cite{xu2020emotion}, without requiring specialized equipment or cooperation from the subject. The main limitation of gait analysis models is the difficulty of gathering properly annotated training data. Images containing faces are prevalent on the internet at large, but gathering an exhaustive gait database requires the willing participation of many subjects to walk in diverse scenarios, especially since gait is affected by many more confounding factors including clothing, carrying conditions and camera viewpoint \cite{Casia}.

Using advantages of both facial and gait analysis, we propose a method to automatically annotate the gender of front-view walking sequences with facial analysis models to generate training data for a gait-based gender recognition system. Through experiments on the popular Front-View Gait (FVG) \cite{FVG} dataset, we show that gender annotations using facial analysis models are on par with ground truth labels, in some cases surpassing facial analysis models, obtaining an F1 score of 91\%. This method is a reliable way to acquire data for training demographic models in real-world scenarios. 
However, due to the limitation of facial analysis models the proposed method only generates annotated data of front-view walking patterns. To overcome this problem, we propose using a semi-supervised approach to generalise to other angles in which the face is not visible. We leverage gait-based metric learning and label propagation to generate noisy pseudo-labels on unseen angles, and use appropriate loss functions developed for training models with label corruption. Training a model on pseudo-labels from label-propagation, we obtain an overall F1 score of 82\% on the CASIA-B \cite{Casia}, similar to a model trained on real labels.

To model walking patterns, we make use of available state-of-the-art pose estimation models (i.e. AlphaPose \cite{li2018crowdpose}), that allow for accurate estimation of the joints of a walking person. We choose to operate on skeleton sequences, as these are more flexible than the standard Gait Energy Images (GEI) used in gait analysis, enabling more fine-grained filtering of the walking sequence, more suitable in real-world scenarios.

While current datasets for modelling demographics from gait (i.e. OULP-Age \cite{OU-ISIR}) are valuable in aiding development of model architectures, we argue that they are less suitable for real-world scenarios, as such controlled scenarios are rarely encountered in surveillance streams. 
Our method is general in the sense that it has the potential to generate annotated data from surveillance streams without human interference. Walking sequences are automatically annotated covering a wide range of real-world scenarios. Through semi-supervised learning, from the annotated front-view walks we obtain annotations for other unseen angles. Models of this pipeline can be replaced with potentially better ones, enabling flexible and general applicability in real-world scenarios.

This paper makes the following contributions:
\begin{itemize}
    \item A method for automatically annotating gender information in front-view walking sequences through facial analysis methods.  We show that models trained on such pseudo-labels have similar performance to facial analysis methods, and surpassing them in adverse scenarios.
    \item A semi-supervised algorithm based on label-propagation to generate pseudo-labels on walking sequences for unseen camera viewpoints, through gait-based metric transfer. Models trained on these pseudo-labels show similar performance to those trained on real labels.
\end{itemize}

%%%%%%%%%%%%%%%%%%%%%%%%%%%%%%%%%%%%%%%%%%%%%%%%%%%%%%%%%%%%%%%%%%%%%%%%%%%%%%%%

% \addtolength{\textheight}{-3cm}   % This command serves to balance the column lengths

%%%%%%%%%%%%%%%%%%%%%%%%%%%%%%%%%%%%%%%%%%%%%%%%%%%%%%%%%%%%%%%%%%%%%%%%%%%%%%%%
\section{RELATED WORK}
In recent years, there has been increased attention towards modelling demographic information from human walking patterns from video. Research in this direction, while still very new, is fuelled by the arrival of new large-scale datasets (i.e. OULP-Age \cite{OU-ISIR}) and the advent of deep learning. Most approaches leverage appearance-based features in the form of a Gait Energy Image (GEI) \cite{gei}. GEIs model the walking sequence be averaging silhouettes across a gait cycle. While this data representation has its limitations by not taking temporal information into account, it is nevertheless successful in applications of gait analysis for both age \cite{age_estimation:Makihara,age_estimation:Sakata, age_estimation:Zhu} and gender \cite{gender_prediction:Lu,Xu_2021_WACV} estimation. 

Notably, Xu et al. \cite{Xu_2021_WACV} proposed a real-time framework for demographic detection from a single silhouette. They firstly reconstruct the gait energy image and process it with a gait recognition network. However, it is unclear if the recognized gender is due to the appearance information of the silhouette or the walking movement patterns. To gauge the impact of movement patterns, skeleton sequences are more appropriate, as they abstract away any appearance features.  However, very few methods exist that make use of skeleton sequences for inferring demographic attributes. Arai et al. \cite{gender-skeleton} inferred the skeleton through morphological manipulation of the silhouette. Ahmed et al. \cite{7985782} used a kinect sensor to estimate the skeleton of the walking subject and analyse gender attributes using handcrafted features on a minimal dataset (18 subjects). In contrast, our goal is to alleviate the need for a carefully gathered dataset of walking sequences and gender, and to develop a pipeline that can generalise to more unconstrained environments. Moreover, as far as the authors know, no method employed skeletons extracted from a modern 2D pose estimation model. Regarding face analysis, many studies have been performed in various conditions, including "in the wild": Zhang et al. \cite{DBLP:journals/corr/abs-1710-02985} developed an architecture called Deep RoR for age and gender identification, that achieves more stable training than other CNN architectures at the time. Rothe et al. \cite{Rothe-IJCV-2018} made a distinction between the real and apparent age of a person, and developed a model that accurately estimates the apparent age of a person based on their face. Significantly more face datasets are available for training and evaluation than gait-based methods. Among the more popular ones include IMDB-WIKI \cite{Rothe-IJCV-2018}, UTKFace \cite{zhifei2017cvpr} and FairFace \cite{DBLP:journals/corr/abs-1908-04913}.

While some methods explore the fusion of face and gait modalities \cite{younis2021hybrid,punyani2018human, gender-face-body}, in which a model jointly learns to identify demographic attributes from both modalities, we instead only use gait for the final prediction, and employ facial traits as a supervisory signal. Our method more closely relates to model distillation \cite{44873}, in which the student model (i.e. gait analysis model) learns to imitate the outputs of the teacher model (i.e. facial analysis models). Moreover, to overcome the limitation of multiple viewpoints, most methods employ a viewpoint classifier, and train an ensemble of models trained on different angles. This assumes labels are present on all viewpoints, which is not reasonable in our case. Instead of using an ensemble, we address this problem using label-propagation guided by gait sequence similarity.

\section{METHOD}
\setlength{\tabcolsep}{0.8em}
{\renewcommand{\arraystretch}{1.1}
\begin{table*}[hbt!]
    \caption{Accuracy of pseudo-labels on CASIA-B, using gait-based label propagation.}
    \label{tab:casia-label-propagation}
    \vspace{-2mm}
    % \resizebox{\linewidth}{!}{
    \centering
    \begin{tabular}{l | ccccccccccc | c}
         &  0$^{\circ}$  &    18$^{\circ}$  &    36$^{\circ}$  &    54$^{\circ}$  &    72$^{\circ}$  &    90$^{\circ}$  &    108$^{\circ}$ &    126$^{\circ}$ &   144$^{\circ}$ &    162$^{\circ}$ &  180$^{\circ}$ & Mean\\
        \hline
        Nearest Neighbors & $-$ &  89.6 &  86.6 &  85.3 &  80.0 &  78.3 &  83.1 &  84.8 &  83.0 &  83.6 & 85.0 & 83.9 \\
        Spectral Clustering & $-$ &  \textbf{92.3} & \textbf{90.5} &\textbf{ 87.0} & \textbf{84.3} &\textbf{ 83.6} & \textbf{85.3} &  \textbf{85.3} &  \textbf{86.1} &  \textbf{85.0} & \textbf{85.5} & \textbf{86.5} \\
        \hline
    \end{tabular}
    % }
    \vspace{-4mm}
\end{table*}
}

Our aim is to develop a method for learning demographic attributes from gait without explicit human annotations. We expand upon the two main components: i) learning to estimate demographics from gait using facial analysis labels on front views (Fig. \ref{fig:framework}) and ii) generalizing to camera viewpoints where the face is not visible using label propagation.

\subsection{Distilling Information from Face Analysis}

For this part of our method, we made use of the Front-View Gait (FVG) dataset \cite{FVG}, a popular dataset for gait recognition from the more challenging front-facing angle. The dataset contains multiple confounding factors such as walking speed (WS), clothing variations (CL), carrying variations (CB) and cluttered background (CBG). The dataset contains high-resolution images, appropriate for facial identification. We repurpose it for gender estimation by manually annotating the 226 subjects with gender information, to have a clear ground truth for comparison.  To analyse facial information, we employ the MTCNN \cite{MTCNN} model to detect faces present in each video, and use an EfficientNet \cite{DBLP:journals/corr/abs-1905-11946} trained on the popular IMDB-WIKI dataset \cite{Rothe-IJCV-2018} to infer demographic attributes. As noted by Gonzalez-Sosa et al. \cite{gender-face-body}, the performance of face-based gender estimation works best on a close distance setting, while gait-based inference is more appropriate at a distance. As such, for final labels we employed a weighted average of the gender confidence based on the area of the face bounding box relative to the frame dimensions. Further, skeletons are extracted for each frame using AlphaPose \cite{li2018crowdpose}, and height normalized. The output of the AlphaPose model consists of 17 joints with X and Y coordinates of the joints and the confidence. To normalize the data, we first zero-center the data by subtracting from each joint the coordinates of the pelvis and then we divide the coordinates of each joint on the X-axis by the length from the right to the left shoulder and on the Y-axis by the length from the pelvis to the neck. Through this process, any appearance information such as the height of the person is removed. In a study by Yang et al. \cite{action_recognition:Yang}, multiple ways of arranging the skeletons are described in order for a model to be able to efficiently learn spatio-temporal information about the movement of a person. We chose the approach in which multiple extracted poses are concatenated to form a skeleton image (TSSI). This way, the spatial information about the joints is preserved, as well as the temporal variation across the gait cycle. We use a ResNet-18 \cite{DBLP:journals/corr/HeZRS15} in all our experiments.

\begin{figure}[hbt!]
    \centering
    \includegraphics[width=0.70\linewidth]{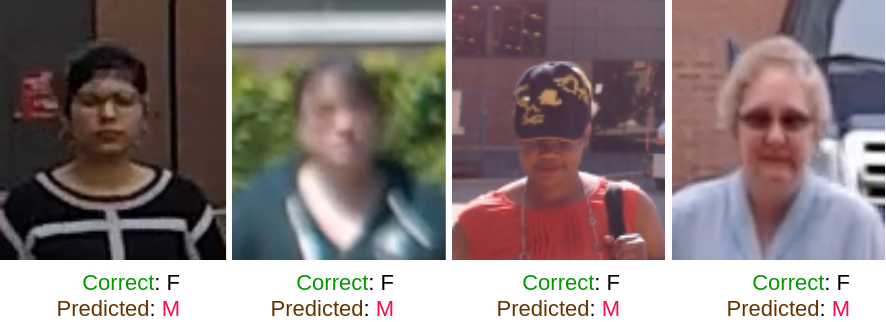}
    \vspace{-2mm}
    \caption{Pseudo-labels generated through facial analysis can be incorrect due to obstructive headware or camera parameters (i.e. camera focus issues).}
    \label{fig:bad-faces}
    \vspace{-2mm}
\end{figure}

Using pseudo-labels generated from facial analysis invariably introduces noise in the training process, as some of these labels can be incorrect due to changes in illumination, head wear etc. Figure \ref{fig:bad-faces} illustrates some of the incorrect labels produced by the facial analysis model. To alleviate the impact of these incorrect labels, we experiment with several methods that handle noisy or incorrect labels, such as more robust training losses \cite{data6060061}, noise modelling (i.e. PENCIL \cite{DBLP:journals/corr/abs-1903-07788} and Importance Reweighting \cite{Liu_2016}). We posit that gait patterns are more robust to extrinsic confounding factors that relate to camera configuration and viewpoint.  

\subsection{Gait-Based Label Propagation}

The previous method is successful at generating pseudo-labels for gender estimation in situations where the face is clearly visible. This corresponds to the majority of cases in which the person is walking directly towards the camera, in front-facing views. One of the major advantages of using gait as an input signal is that is has the potential to work in more adverse situations than facial analysis systems, mainly in scenarios in which the subject is far from the camera or not facing towards it. However, by only using demographic attributes estimated from front-facing views, the gait analysis model is unable to generalise to other angles. To solve this issue, we propose to use a method similar to that of Liu et al. \cite{DBLP:journals/corr/abs-1812-08781}. The authors propose an algorithm for deep label propagation, using a pretrained metric model. Given a small set of labelled examples, labels are propagated to the large number of unlabelled samples by making use of a semantic similarity measure between examples. In our case, we consider only the front-facing views to be labelled and other viewpoints in which the face is not visible to be unlabelled. However, different from the simple case of images, walking sequences from different viewpoints are qualitatively very different from one another. Previous work in unsupervised gait recognition by Cosma et al. \cite{cosma2021wildgait} showed that a similarity metric between walking sequences can be effectively learned without human labels, by leveraging contrastive learning and massive amounts of unlabelled data. WildGait, the framework proposed by Cosma et al. \cite{cosma2021wildgait} for gait-based person identification using skeletons, achieves disentanglement of viewpoint variation and confounding factors, and is a suitable candidate to be used as a similarity measure between walking sequences. As such, we leverage a pretrained WildGait network as a similarity network between walking sequences, and use it to further propagate labels to unseen angles.

As FVG does not contain viewpoints in which the face is not visible, we conduct experiments on CASIA-B, which offers 11 different viewpoints. We manually annotate subjects with gender attributes to have an accurate benchmark. 

\setlength{\tabcolsep}{0.8em}
{\renewcommand{\arraystretch}{1.1}
\begin{table*}[hbt!]
    \caption{Results on gender classification on CASIA-B using only frontview gender annotations (i.e. 0$^{\circ}$).}
    \label{tab:casia-semisupervised}
    \vspace{-2mm}
    \resizebox{\textwidth}{!}{
    \centering
    \begin{tabular}{l | ccccccccccc | c }
        & \multicolumn{12}{ c}{}\\
         &  0$^{\circ}$  &    18$^{\circ}$  &    36$^{\circ}$  &    54$^{\circ}$  &    72$^{\circ}$  &    90$^{\circ}$  &    108$^{\circ}$ &    126$^{\circ}$ &   144$^{\circ}$ &    162$^{\circ}$ &    180$^{\circ}$ & Mean \\
        \hline
        Baseline (train only on 0$^{\circ}$) & 69.92 &   33.33 &   33.33 &   33.33 &   33.33 &   33.33 &    46.89 &    31.03 &    38.91 &    72.34 &    \textbf{84.85} &  46.41 \\
        \hline
        Gait-Based Label Propagation - NN & 76.30 &   85.0 &   \textbf{65.77} &   \textbf{87.3} &   52.38 &   56.36 &    67.03 &    70.25 &    71.63 &    70.25 &    81.95 &  71.29 \\
        Gait-Based Label Propagation - Spectral & \textbf{89.90} &   \textbf{97.50} &   63.66 &   33.33 &   \textbf{97.50} &  \textbf{ 89.90} &   \textbf{ 92.46} &    \textbf{89.90} &   \textbf{ 76.30} &   \textbf{ 92.46} &    81.95 &  \textbf{82.26} \\
        \hline
        True Labels (all viewpoints) & 79.80 &   92.50 &   80.0 &   73.33 &  100.0 &   97.50 &   100.0 &    94.99 &    79.17 &    92.46 &    84.65 &  88.58 \\
        \hline
    \end{tabular}
    }
    \vspace{-2mm}
\end{table*}
}

\section{EXPERIMENTS \& RESULTS}
\subsection{Distilling Information from Face Analysis: FVG}

To avoid a favorable configuration of our setting, we used cross-validation by randomly sampling 46 unique subjects and utilizing their data for validation while the remaining identities are used for training. We ran each experiment three times. Since FVG contains 64\% male subjects and 36\% female subjects, we oversampled from the female class during the training process in order to overcome class imbalance. Due to the disparity in the validation data, we report the mean F1 score and standard deviations. In all experiments we used ResNet-18 \cite{DBLP:journals/corr/HeZRS15} architecture, Adam optimizer \cite{Adam} with a learning rate of 0.0001 and batch size of 128. We employed the standard array of augmentations for gait analysis: skeleton flipping, mirroring, joint dropout, pace modification and random cropping.

\begin{figure}[hbt!]
    \centering
    \includegraphics[width=0.70\linewidth]{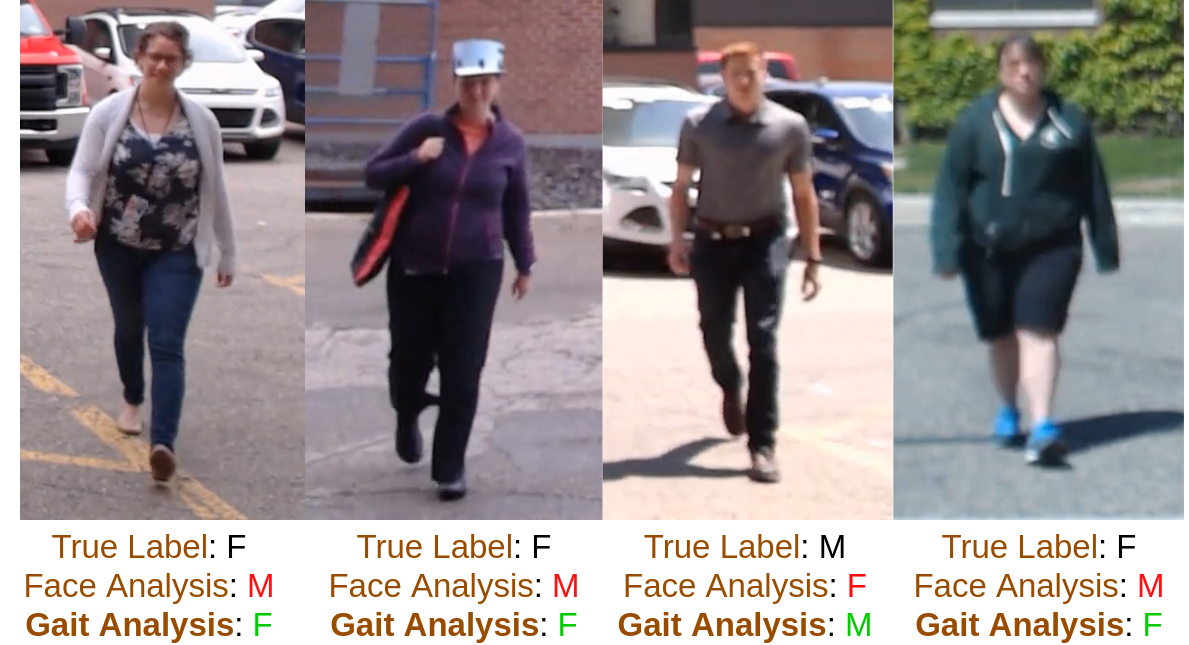}
    \caption{Selected samples in which facial analysis outputs the wrong gender due to extrinsic camera parameters, to which the gait model is robust.}
    \label{fig:face-bad-gait-good}
    \vspace{-5mm}
\end{figure}
Since pseudo-labels can be noisy, we compare several methods that account for incorrect labels: Probabilistic End-to-end Noise Correction (PENCIL) \cite{DBLP:journals/corr/abs-1903-07788}, a method agnostic to the backbone network and any prior knowledge of the noise distribution; Importance Reweighting \cite{Liu_2016}, a method widely used in domain adaptation; a framework proposed Ciortan et al. \cite{data6060061} for classification with noisy labels with promising results using a loss function designed for adversarial robustness: a combination of Negative Focal Loss and Reverse Cross Entropy (NFL-RCE).

Table \ref{tab:results-noise} shows our results in terms of F1 score on multiple walking variations from FVG. The gait-based model trained on real gender annotations has good performance on the FVG test set, similar to the face-based model. This is an encouraging result in terms of the reliability of gait-based gender prediction systems.
Since the gait-based model trained on real gender labels or pseudo-labels has similar, good results demonstrates the effectiveness of our method of training gait-based models with pseudo-labels.
Also, a model that relies on face analysis has shortcomings that gait analysis can overcome. Figure \ref{fig:face-bad-gait-good} displays scenarios in which a gait-based model trained on pseudo-labels produces the correct output where the face analysis model is incorrect. Blurry faces or people wearing hats or glasses tend to make the face analysis model underperform. In comparison, gait analysis is more robust to extrinsic confounding factors. 
These results highlight the efficiency of our method. Even though the gait-based model is trained on labels generated by face analysis, in some difficult cases for face analysis (such as the ones in Figure \ref{fig:face-bad-gait-good}), it is able to generalise despite having been provided with unreliable labels.
Moreover, the addition of methods that incorporate noisy labels further improves the classification robustness.

\setlength{\tabcolsep}{0.8em}
{\renewcommand{\arraystretch}{1.1}
\begin{table}[hbt!]
    \caption{F1 score of the gait analysis model on FVG trained with different methods that account of noisy training labels.}
    \label{tab:results-noise}
    \vspace{-2mm}
    \resizebox{\linewidth}{!}{
    \centering
    \begin{tabular}{|c|c c c c c|}
    \hline
      & WS & CB & CL & CBG & All \\ 
     \hline\hline
     Face & 91.84 $\pm$ 3.1 & 89.14 $\pm$ 0.7 & 87.6 $\pm$ 6.8 & 91.7 $\pm$ 6.0 & 91.0 $\pm$ 2.7 \\
    \hline
     Gait  True Labels & 92.0 $\pm$ 3.1 & 87.2 $\pm$ 2.8 & 92.7 $\pm$ 7.0 & 91.0 $\pm$ 8.3 & 92.0 $\pm$ 2.6 \\
     \hline
     Gait  Face Labels &\textbf{ 92.0 $\pm$ 1.7} & 82.7 $\pm$ 8.3 & 91.0 $\pm$ 1.1 & 86.2 $\pm$ 5.7 & 90.7 $\pm$ 1.9 \\
     NFL-RCE \cite{data6060061} & 88.6 $\pm$ 2.7 & \textbf{86.1 $\pm$ 4.9} & 83.5 $\pm$ 2.5 & 83.7 $\pm$ 8.4 & 88.9 $\pm$ 1.7 \\
     IW$l$ \cite{Liu_2016} & 90.4 $\pm$ 0.5 & 82.2 $\pm$ 8.8 &\textbf{ 95.3 $\pm$ 3.3} & 89.9 $\pm$ 7.8 & 90.1 $\pm$ 1.2 \\
     PENCIL \cite{DBLP:journals/corr/abs-1903-07788} & 90.8 $\pm$ 2.3 & 85.0 $\pm$ 7.1 & 92.1 $\pm$ 0.8 &\textbf{ 90.6 $\pm$ 7.5} & \textbf{91.0 $\pm$ 1.7} \\
     \hline
    \end{tabular}
    }
    \vspace{-4mm}
\end{table}
}

\subsection{Gait-Based Label Propagation: CASIA-B}

To evaluate our method in scenarios where only frontview walks are annotated with gender labels, we utilize the popular CASIA-B \cite{Casia} dataset for gait recognition. CASIA-B contains 124 subjects walking from 11 different viewpoints, with clothing variations and carrying conditions. The dataset was manually annotated with gender labels, and 92 males and 32 females were obtained. The data was split into 80\% training and 20\% validation and the majority class (males) was undersampled to balance the dataset.

We follow the Deep Metric Transfer framework of Liu et al. \cite{DBLP:journals/corr/abs-1812-08781}, in which label propagation is performed using a pretrained metric model between samples. In our case, we employed WildGait \cite{cosma2021wildgait} as a similarity measure between walking sequences. WildGait was trained on UWG, a large-scale dataset of noisily annotated walking sequences from real-world scenes, in an unsupervised manner. The model was not fine-tuned on CASIA-B and was used as-is.

We considered only the frontview angle (0$^\circ$) to be annotated with gender labels, and employed label-propagation to annotate other angles in which the face is not visible. We compared 2 label-propagation methods: nearest neighbors and spectral clustering. Similar to Liu et al. \cite{DBLP:journals/corr/abs-1812-08781}, best results were obtained with the latter, as shown in Table \ref{tab:casia-label-propagation}.

Further, we trained a ResNet-18 on the propagated pseudo-labels, together with the true labels at angle 0$^\circ$. We employed the NFL-RCE loss function to account for incorrect pseudo-labels. F1 scores are presented in Table \ref{tab:casia-semisupervised} for all angles in CASIA-B. Compared to the baseline model, label-propagation achieves significantly better results. Moreover, spectral label propagation is similar in performance to the model trained on data with all available true labels, highlighting the capability of label propagation using a gait-based similarity measure to generalise to unseen angles.

%%%%%%%%%%%%%%%%%%%%%%%%%%%%%%%%%%%%%%%%%%%%%%%%%%%%%%%%%%%%%%%%%%%%%%%%%%%%%%%%
\section{CONCLUSIONS}
This work proposes and evaluates a system to estimate the gender of a person based on their walking patterns, without requiring any explicitly labelled data. 
We make use of current state-of-the-art facial analysis systems to automatically annotate videos of walking persons. However, facial analysis models have two inherent problems: i) outputs are not always correct and ii) only work on walking from front-views. To overcome these shortcomings, we employed several methods to account for noisy labels, which increases the robustness of our model, and employed a gait-based label propagation method to generalise to unseen walking angles.

The results show great potential of automatic labeling of gait patterns to be used in practice in surveillance scenarios, with performance on par with facial analysis models, and oftentimes exceeding it, as gait analysis models are inherently more robust to extrinsic confounding factors, such as camera configuration and walking viewpoint.

%{\color{red} Any work involving the estimation of gender from biometric information has limitations regarding the issues of gender identity. Our facial analysis model was trained on IMDB-WIKI, and has the same limitations as existing facial analysis models that use a binary model for describing gender.}
%%%%%%%%%%%%%%%%%%%%%%%%%%%%%%%%%%%%%%%%%%%%%%%%%%%%%%%%%%%%%%%%%%%%%%%%%%%%%%%%

{\small
\bibliographystyle{ieee}
\bibliography{refs}
}

\end{document}